\PassOptionsToPackage{table,xcdraw,dvipsnames}{xcolor}
\documentclass[acmsmall,nonacm,screen,11pt]{acmart}
\setcopyright{none}
\acmYear{2026}
\usepackage{brand}
\usepackage{header}
\usepackage{docspec}
\renewfontfamily\AppliedDisplayBold{AppliedSansDisplay-Bold}[
  Path=fonts/, Extension=.ttf, BoldFont=AppliedSansDisplay-Bold]

\usepackage{nicefrac}
\usepackage{needspace,threeparttable,xspace,multirow,nccmath,adjustbox}
\usepackage{ragged2e,wrapfig,appendix,makecell}
\usepackage[ruled,vlined,linesnumbered]{algorithm2e}
\def\thename{LaMo\xspace}
\newcommand{\boldparagraph}[1]{\par\noindent\textbf{#1}}
\renewcommand{\paperurl}{https://lamo-ai.github.io/}

\begin{document}
\title[LaMo: Latent Motion Priors]{LaMo: Self-Supervised Latent Motion Priors\\for Physical Realism in Video Generation}

\author{Bo Jiang}
\affiliation{\institution{Applied Intuition}\city{Mountain View}\state{CA}\country{USA}}
\author{Depu Meng}
\affiliation{\institution{Applied Intuition}\city{Mountain View}\state{CA}\country{USA}}
\author{Yihan Hu}
\affiliation{\institution{Applied Intuition}\city{Mountain View}\state{CA}\country{USA}}
\author{Yichen Xie}
\affiliation{\institution{Applied Intuition}\city{Mountain View}\state{CA}\country{USA}}
\affiliation{\institution{University of California, Berkeley}\city{Berkeley}\state{CA}\country{USA}}
\author{Tianshuo Xu}
\affiliation{\institution{Applied Intuition}\city{Mountain View}\state{CA}\country{USA}}
\author{Wei Zhan}
\affiliation{\institution{Applied Intuition}\city{Mountain View}\state{CA}\country{USA}}
\affiliation{\institution{University of California, Berkeley}\city{Berkeley}\state{CA}\country{USA}}

\compactauthors{%
  {\raggedright
   {\rmfamily\bfseries
    Bo Jiang\textsuperscript{1,*}, \;
    Depu Meng\textsuperscript{1}, \;
    Yihan Hu\textsuperscript{1}, \;
    Yichen Xie\textsuperscript{1,2,*}, \;
    Tianshuo Xu\textsuperscript{1,*}, \;
    Wei Zhan\textsuperscript{1,2,\dagger}\par}
   \smallskip
   {\small\rmfamily\color{AppliedGray}%
    \textsuperscript{1}Applied Intuition \quad
    \textsuperscript{2}University of California, Berkeley\par}
   \smallskip
   {\footnotesize\rmfamily\color{AppliedGray}%
    \textsuperscript{*}Work done during internship at Applied Intuition.\\\textsuperscript{\dagger}Corresponding author: \texttt{wei.zhan@applied.co}.\par}}
}

\begin{teaserfigure}
\begingroup
\renewenvironment{figure}[1][]{}{}
\begin{figure}[t]
    \centering
    \includegraphics[width=0.99\linewidth]{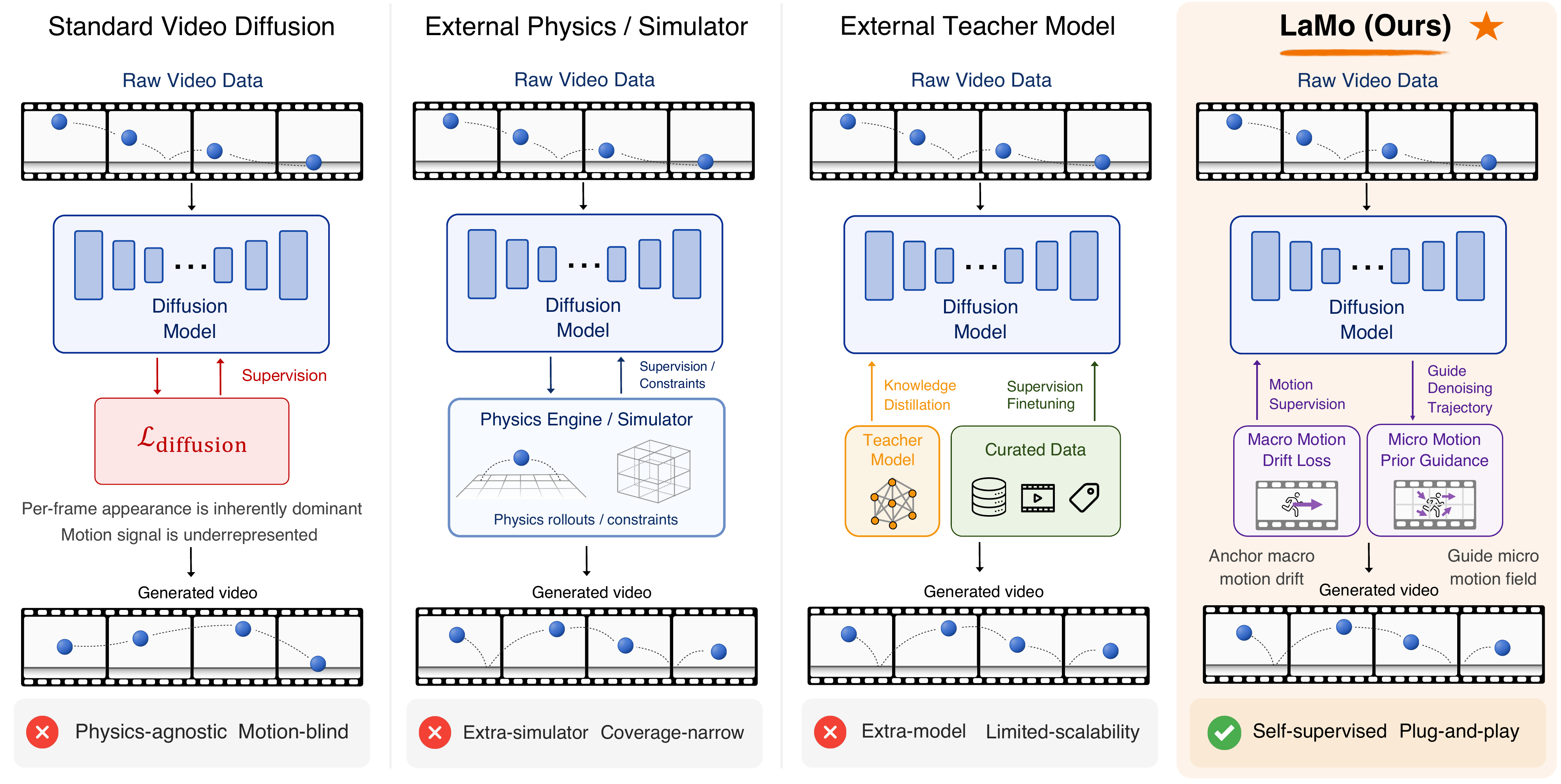}
    \caption{Existing approaches to physical realism rely on hand-crafted physics simulators (Col.~2) or on foundation-model teachers and curated physics-annotated corpora (Col.~3), while the vanilla diffusion objective is dominated by per-frame appearance (Col.~1). \thename{} (Col.~4) instead extracts motion supervision from the same raw video data through a Motion Drift Loss during training and Motion Prior Guidance during inference.}
    \label{fig:teaser}
\end{figure}

\endgroup
\end{teaserfigure}

\begin{abstract}
Modern video generators produce visually compelling clips but still struggle with physical and motion consistency, limiting their use as reliable world simulators. Existing remedies often rely on external simulators, teacher models, or curated physics-focused data. We explore a complementary self-supervised direction: extracting motion cues from the unlabeled videos already used to train video diffusion models. We propose \thename{}, which formulates a latent motion prior over frame-to-frame latent changes conditioned on the current latent and prompt. This prior is exposed through two lightweight readouts: a macro motion drift used during training as a Motion Drift Loss, and a learned micro motion field used during sampling as Motion Prior Guidance. Both components are plug-and-play with existing video diffusion backbones, requiring no architectural or I/O changes. On VideoPhy, VideoPhy2, and PisaBench, \thename{} improves CogVideoX backbones and outperforms recent physics-aware baselines that use external supervision. On VBench, it preserves overall generation quality while improving motion-related dimensions. These results suggest that unlabeled video contains useful motion supervision for improving physical fidelity in modern video diffusion models.
\end{abstract}

\maketitle
\hypersetup{pdftitle={LaMo: Self-Supervised Latent Motion Priors for Physical Realism in Video Generation},pdfauthor={Bo Jiang, Depu Meng, Yihan Hu, Yichen Xie, Tianshuo Xu, Wei Zhan}}
\clearpage
\section{Introduction}
\label{sec:intro}

Video generation has advanced rapidly in recent years, with modern diffusion- and transformer-based models producing high-resolution, temporally coherent, and visually compelling videos~\citep{wiedemer2025video, wan2025wan, gao2025seedance}. This progress has motivated a broader view of video generators as potential world simulators for spatial reasoning, embodied planning, and synthetic data generation~\citep{agarwal2025cosmos, bar2025navigation, hafner2025dreamerv3}. However, visual realism alone is insufficient for this goal: a useful world simulator must also capture how the physical world evolves over time, making physical and motion fidelity a central challenge for modern video generation.

A persistent limitation of current video generators is that they often produce motion that appears plausible locally but violates basic physical regularities, such as gravity, contact, material interaction, or deformation consistency~\citep{bansal2024videophy, bansal2025videophy2, wu2026motion}. This suggests a mismatch between the standard training objective and the desired notion of physical realism. The diffusion denoising loss supervises per-token reconstruction, but it does not explicitly target the frame-to-frame changes that define motion and physical evolution. As a result, temporal coherence can improve with scale, yet physically consistent dynamics remain difficult to guarantee.

Recent methods address this issue by introducing additional sources of physical knowledge. Some approaches couple video generation with simulators or explicit dynamics models~\citep{liu2026realwonder, foo2026pisvg, yuan2025newtongen}, while others rely on physics-aware rewards, curated physics datasets, or distillation from auxiliary foundation models~\citep{tan2026dreamworld, zhang2025videorepa, wang2025wisa}. These methods have shown promising improvements, but they also introduce external dependencies, domain-specific assumptions, or additional supervision. In contrast, large-scale unlabeled video already contains rich evidence of how objects move, interact, and deform. This motivates a simpler question: \textit{can the motion signal already present in ordinary video data be extracted and reused to improve physical fidelity in video diffusion models?}

We propose \thename{} (Latent Motion Prior), a self-supervised framework that models physical and motion regularities through frame-to-frame latent changes. Given clean video latents $z$, we define the $\tau$-step latent difference $\Delta_\tau z := z^{(i+\tau)} - z^{(i)}$ and formulate a latent motion prior $p(\Delta_\tau z \mid z, c)$, where $c$ denotes the conditioning already used by the backbone. This prior requires no physical annotations, simulators, or teacher models: every pair of nearby frames in an unlabeled video provides a training signal.

To make this prior usable in existing diffusion pipelines, \thename{} exposes it through two lightweight readouts. First, a macro motion drift captures the channel-wise rate of latent change and is used during training as a Motion Drift Loss. Second, a learned micro motion field predicts spatially resolved latent changes and is used during sampling as Motion Prior Guidance. Both components are designed to preserve the original backbone architecture and input/output interface.

Experiments on VideoPhy~\citep{bansal2024videophy}, VideoPhy2~\citep{bansal2025videophy2}, and VBench~\citep{huang2024vbench} show that \thename{} improves physical and motion fidelity while maintaining general generation quality. PisaBench~\citep{li2025pisabench} adds a reference-based object-motion evaluation. Compared with recent physics-aware methods that use external supervision, \thename{} achieves competitive or stronger results using only self-supervised signals from unlabeled video. These results suggest that latent motion differences provide a useful and minimally invasive source of supervision for improving physical realism in modern video diffusion models.

Our contributions are summarized as follows:

\begin{itemize}[leftmargin=12pt]

  \item \textbf{Self-supervised latent motion prior.} We formulate physical and motion regularities as a latent motion prior over frame-to-frame latent changes, which can be estimated directly from unlabeled video without simulators, physical annotations, or teacher models.

  \item \textbf{Lightweight training and sampling readouts.} We introduce a macro Motion Drift Loss for training and a micro Motion Prior Guidance for sampling, both of which preserve the backbone architecture and input/output interface.

  \item \textbf{Improved physical fidelity with preserved general video quality.} On VideoPhy and VideoPhy2, \thename{} improves CogVideoX backbones and is competitive with recent externally supervised physics-aware baselines; on VBench, it maintains overall video generation quality while improving motion-related metrics. PisaBench provides complementary object-motion evidence.

\end{itemize}

\section{Related Work}
\label{sec:related}

\boldparagraph{Physical and Motion Fidelity in Video Generation.}
Recent diffusion- and transformer-based video generators have advanced substantially in text alignment, visual quality, and temporal coherence~\citep{yang2024cogvideox,kong2024hunyuanvideo,wan2025wan,gao2025seedance}. Yet high visual fidelity does not guarantee physically faithful dynamics: physics-oriented evaluations show frequent violations of gravity, contact, material interaction, deformation, and conservation-related commonsense~\citep{bansal2024videophy,bansal2025videophy2}. Motion-focused analyses similarly show that clips improving appearance need not improve temporal dynamics~\citep{wu2026motion}. These findings make physical and motion fidelity a central bottleneck for reliable world simulation, which requires plausible temporal evolution rather than only plausible frames.

\boldparagraph{Physics-Aware Video Generation.}
Existing attempts mostly inject additional physical or motion knowledge into the generator. One line grounds generation in explicit priors, using simulator-produced trajectories, reconstructed physical properties, physical simulation, dynamical equations, or learnable Newtonian dynamics as conditioning or generation signals~\citep{wang2025physctrl,zhang2025physchoreo,liu2026realwonder,foo2026pisvg,yuan2025newtongen}. Motion-aware objectives or inference-time guidance further compensate for appearance-oriented training objectives that under-emphasize motion coherence~\citep{chefer2025videojam}. These methods provide controllability for targeted phenomena, but reliance on solvers, material parameters, 3D reconstruction, scene-specific states, or specialized motion representations limits scalability to unconstrained videos, especially when such signals must be estimated before generation.

A second line supplies physical knowledge through post-training objectives, curated data, or auxiliary foundation models. Reward- and preference-based methods enforce Newtonian constraints, collision rules, force-conditioned goals, VLM-derived preferences, or 3D consistency through RL-style fine-tuning~\citep{le2025NewtonReward,zhang2026physrvg,gillman2026goalforce,cai2025phygdpo,wang2026worldr1}. WISA decomposes physical principles into textual, qualitative, and quantitative descriptors and injects them through Mixture-of-Physical-Experts attention trained on curated physics data~\citep{wang2025wisa}. VideoREPA distills token-level physical relations from video foundation models into text-to-video diffusion transformers~\citep{zhang2025videorepa}. Other efforts align generation with VLM/VFM-derived physical cues, latent physical dynamics, or local physics annotations~\citep{wang2025prophy,shen2026phantom,satish2026physvideogenerator,pathak2026physvid}. These methods are effective but still depend on external teachers, annotations, handcrafted rewards, simulator rollouts, or architectural components. In all cases, the physical signal is imported from an auxiliary source; we instead extract supervision from ordinary unlabeled videos already consumed by the generator.

\boldparagraph{Self-Supervised Video Dynamics.}
Self-supervised video learning provides evidence that raw video contains rich dynamics supervision. Masked reconstruction methods such as VideoMAE learn transferable video representations from unlabeled clips, but primarily target recognition and representation transfer~\citep{tong2022videomae,wang2023videomaev2}. Latent predictive models offer a more direct connection to dynamics: V-JEPA, V-JEPA~2, and LeWorldModel predict masked spatio-temporal regions or compact latent states while encoding physical structure~\citep{bardes2024vjepa,assran2025vjepa2,maes2026leworldmodel}. Probing and counterfactual world-modeling studies further show that latent predictors can acquire intuitive physics and dynamics-relevant structures such as keypoints, flow, and segmentations~\citep{garrido2025intuitive,venkatesh2024understanding}. Interactive world models such as Genie and VideoWorld demonstrate that unlabeled videos can support control, planning, and transferable task knowledge~\citep{bruce2024genie,ren2025videoworld,ren2026videoworld2}. However, these works mainly target representation learning, planning, or interactive rollout, so their dynamics knowledge is not directly consumed by high-fidelity text-to-video diffusion models; \thename{} converts raw-video dynamics into direct training and sampling signals.

\begin{figure}[t]
    \centering
    \includegraphics[width=0.99\linewidth]{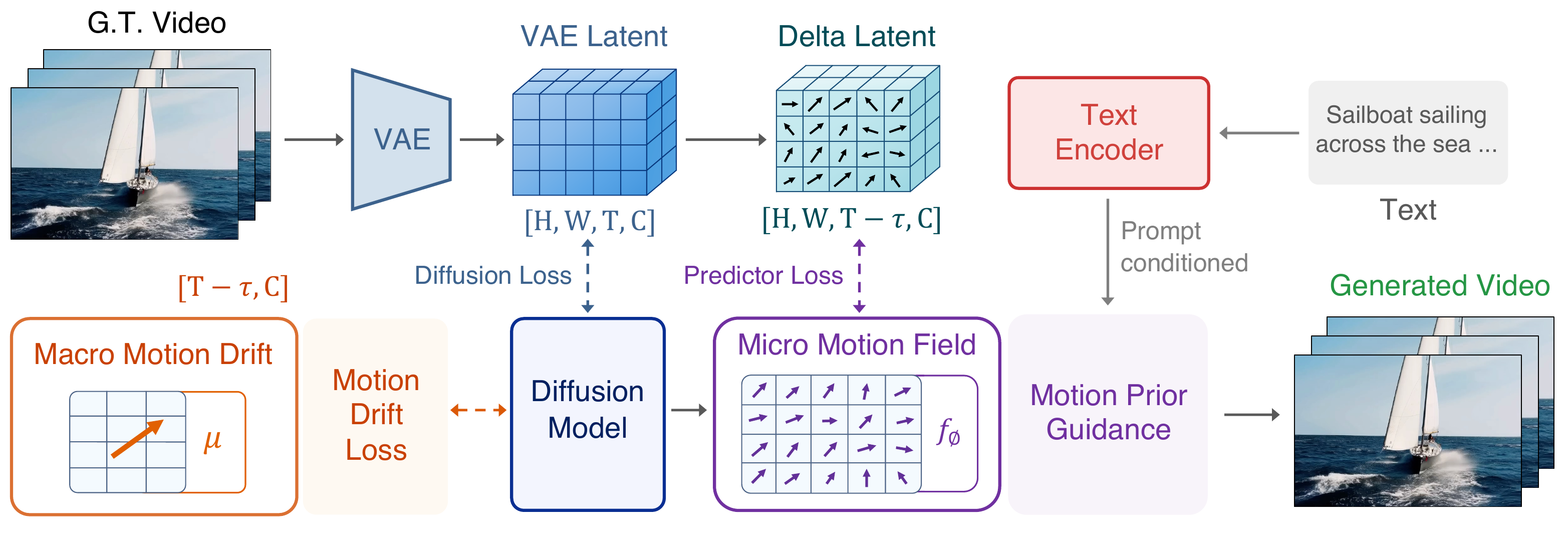}
    \caption{\textbf{\thename{} Overview.} Frame-to-frame differences of clean VAE latents yield a self-supervised latent motion prior, exposed through two readouts: a \emph{macro motion drift} used as a training-time \emph{Motion Drift Loss}, and a lightweight learned \emph{micro motion field} used at sampling as \emph{Motion Prior Guidance}. Both share the backbone's prompt conditioning and leave its architecture unchanged.}
    \label{fig:method}
\end{figure}

\section{Method}
\label{sec:method}

We organize \thename{} around the method overview's data flow. First, §\ref{sec:method:prelim} encodes raw videos into clean video latents and defines the latent motion prior over frame-to-frame latent changes. Second, §\ref{sec:method:bmv} extracts the prior's stable macro component and uses it as a training-time \emph{Motion Drift Loss}. Third, §\ref{sec:method:guide} learns a spatially resolved micro motion field and applies it as sampling-time \emph{Motion Prior Guidance}. Finally, §\ref{sec:method:pipeline} summarizes the full training and inference pipeline, including the added cost.

\subsection{Problem Setup and Latent Motion Encoding}
\label{sec:method:prelim}

\boldparagraph{Notation.}
Let $x$ be a video clip and $z_{1:T}=\mathcal{E}(x)\in\mathbb{R}^{T\times C\times H\times W}$ its latent sequence encoded by a pretrained 3D causal VAE $\mathcal{E}$; $T$ counts latent frames after temporal compression. We use $i \in \{1, \dots, T\}$ for the \emph{frame index} and $t \in \{1, \dots, S\}$ for the \emph{diffusion timestep}; $\bar\alpha_t \in (0, 1]$ is the cumulative noise schedule and $\sigma_t := \sqrt{1 - \bar\alpha_t}$. We write $c$ for any auxiliary conditioning the backbone already accepts. The diffusion forward process produces $z_t^{(i)} = \sqrt{\bar\alpha_t}\,z_0^{(i)} + \sigma_t\,\epsilon^{(i)}$, and the backbone is trained with the standard per-token denoising loss $\mathcal{L}_\mathrm{denoise} = \mathbb{E}_{t, z_0, \epsilon}\big[\|\epsilon_\theta(z_t, t, c) - \epsilon\|_2^2\big]$. We write $\hat x_0(\hat\epsilon, z_t, t)$ for the data-side projection of the model state, recoverable in closed form under either $\epsilon$- or $v$-prediction.

\boldparagraph{The latent change object.}
Physical and motion realism is a property of \emph{how} latents change between frames, not of \emph{what} each latent contains. We therefore take as the carrier of dynamics the $\tau$-step latent change on clean latents,
\begin{equation}
  \Delta_\tau z^{(i)} \;:=\; z_0^{(i+\tau)} - z_0^{(i)} \;\in\; \mathbb{R}^{C \times H \times W},
  \qquad i \in \{1, \dots, T - \tau\}.
\end{equation}
The choice $\tau \in \mathbb{N}$ sets the temporal scale: smaller values capture shorter-range changes, while larger values emphasize lower-frequency drift.

\boldparagraph{Latent motion prior.}
The standard denoising objective remains the main supervision for appearance and prompt-conditioned video synthesis, but it does not explicitly supervise the clean latent increment $\Delta_\tau z$ that carries frame-to-frame motion. We therefore collect the motion signal available in unlabeled videos into a conditional prior,
\begin{equation}
  p(\Delta_\tau z \mid z, c),
\end{equation}
where the conditioning variables are already present in the diffusion pipeline. This prior is self-supervised: every pair of nearby frames in an ordinary training clip provides a sample of $\Delta_\tau z$, requiring no simulator state, physics annotation, or external teacher.

\boldparagraph{Macro and micro readouts.}
\label{sec:method:readouts}
We use the first moment of the prior,
\begin{equation}
  m(z,c) := \mathbb{E}[\Delta_\tau z \mid z,c] \in \mathbb{R}^{C \times H \times W},
\end{equation}
as the operational target of \thename{}. It naturally separates into a spatially averaged macro component and a spatially resolved micro component. The macro drift is:
\begin{equation}
  \mu(z,c) := \mathbb{E}_{H,W}[m(z,c)] \in \mathbb{R}^{C},
\end{equation}
broadcastable over $H \times W$, while the residual field describes where and how this drift is allocated across the scene. This split gives two complementary readouts:

\begin{itemize}[leftmargin=12pt]
  \item \textbf{Macro readout (motion drift).} The channel-wise drift $\mu(z, c)$ admits a parameter-free closed-form estimator on any minibatch of clean clips, $\hat\mu = \frac{1}{HW}\sum_{h,w}\Delta_\tau z$. No predictor is fit; the observed statistic supplies a self-supervised target for the prior's first moment at this granularity.
  \item \textbf{Micro readout (motion field).} The full first moment is high-dimensional, so we model it by a small predictor $f_\phi(z, c) \approx m(z, c)$ trained on raw latent pairs.
\end{itemize}

Although both readouts come from the same prior, their granularity and estimation stability make them better suited to different stages of the diffusion pipeline. The macro motion drift is parameter-free and stable at minibatch scale, making it a natural auxiliary target during backbone fine-tuning. At inference, however, it would provide only a spatially uniform correction, i.e., a single vector in $\mathbb{R}^C$ broadcast over $H \times W$, and therefore cannot indicate where motion should occur. Conversely, the learned micro field captures spatially resolved motion, but using it as a training-time target would introduce an additional source of co-evolving supervision whose errors may interact with backbone optimization. We therefore use the macro readout as a training-time loss (§\ref{sec:method:bmv}) and the micro readout as an inference-time guidance term (§\ref{sec:method:guide}), where the predictor is frozen and only steers the sampling trajectory. This stage-matched design is supported empirically in §\ref{sec:exp:ablation}: replacing the channel-mean drift target with dense per-pixel motion supervision degrades both physical-fidelity axes relative to our method (Table~\ref{tab:design_ablation}).

\subsection{Macro Readout: Latent Motion Drift Loss}
\label{sec:method:bmv}

We now derive the training-time \textbf{Motion Drift Loss} from the macro drift $\mu$.

\boldparagraph{Anchor on $\hat x_0$.}
We anchor the loss on $\hat x_0$, the data-side projection of the model state and the very target that $\mathcal{L}_\mathrm{denoise}$ optimizes, so that the auxiliary term composes additively without rescaling between timesteps; an analogous constraint on $\hat\epsilon$ would change units between steps, while one placed on $z_t$ would not depend on $\theta$ at all. The same anchor carries over to inference (§\ref{sec:method:guide}) with the differentiation variable switched from $\theta$ to $\hat\epsilon$, a deliberate symmetry between the two stages. We write $\hat\mu^{(i)} := \mathbb{E}_{H,W}[\hat x_0^{(i+\tau)} - \hat x_0^{(i)}]$ and $\mu^{\star,(i)} := \mathbb{E}_{H,W}[z_0^{(i+\tau)} - z_0^{(i)}]$ for the predicted and ground-truth drifts at frame index $i$.

\boldparagraph{Lag choice.}
The lag $\tau$ controls the temporal bandwidth of the supervision. We use $\tau{=}2$ by default: in the two-point VideoPhy ablation, it outperforms $\tau{=}1$ (Table~\ref{tab:design_ablation}). This is consistent with a possible contribution from short-range VAE jitter, but the ablation does not isolate that mechanism or compare latent with pixel-space targets.

\boldparagraph{Scale-normalized L2.}
We seek a distance that decouples \emph{magnitude} from \emph{direction} error without introducing a free trade-off hyperparameter. The scale-normalized form:
\begin{equation}
  \mathcal{L}_\mathrm{drift} \;=\; \mathbb{E}_{b, i}\bigg[\frac{\|\hat\mu^{(i)} - \mu^{\star,(i)}\|_2^2}{\mathrm{sg}\big[\|\mu^{\star,(i)}\|_2^2 + \varepsilon\big]}\bigg]
  \;\approx\; (r - 1)^2 \;+\; 2\,r\,(1 - \cos\vartheta),
  \label{eq:drift}
\end{equation}
with $r := \|\hat\mu\|/\|\mu^\star\|$ and $\vartheta := \angle(\hat\mu, \mu^\star)$, reduces to the right-hand decomposition when $\varepsilon$ is negligible. The small constant $\varepsilon$ prevents quiet clips from dominating the gradient when $\|\mu^\star\|$ is close to zero, while the stop-gradient $\mathrm{sg}[\cdot]$ keeps the denominator as a per-clip rescaling factor rather than a learnable target.

\boldparagraph{Final training objective.}
At high noise levels $\hat x_0$ is dominated by $\hat\epsilon$ and is uninformative about the data drift, so we damp the loss with a per-batch averaged schedule weight $w(\sigma_t) := \mathbb{E}_b[\bar\alpha_{t,b}]$ that vanishes at maximal noise. The final training objective is:
\begin{equation}
  \mathcal{L}_\mathrm{train} \;=\; \mathcal{L}_\mathrm{denoise} \;+\; \lambda_\mathrm{drift}\,\cdot\,w(\sigma_t)\,\cdot\,\mathcal{L}_\mathrm{drift},
  \label{eq:train}
\end{equation}
which is parameter-free apart from the scalar $\lambda_\mathrm{drift}$ and the numerical stabilizer $\varepsilon$. The schedule weight suppresses the auxiliary loss in the high-noise regime where $\hat x_0$ is unreliable, while the normalized denominator prevents low-motion clips from producing unstable gradients. By construction, $\mathcal{L}_\mathrm{drift}$ constrains only the macro rate of latent change; the spatial allocation of motion is handled by the inference-time micro readout.

\subsection{Micro Readout: Motion Prior Guidance}
\label{sec:method:guide}

The micro readout $f_\phi$ is consumed at sampling time as a CFG-style~\citep{ho2022classifier} gradient on the noise prediction. We first describe the predictor itself, then justify why this signal is best applied in noise-prediction space rather than directly on the latent.

\boldparagraph{Predictor and training.}
$f_\phi : \mathbb{R}^{C \times H \times W} \times \mathbb{R}^{D_c} \to \mathbb{R}^{C \times H \times W}$ is a small CNN ($\sim$10M parameters, against $\sim$2--5B for the backbones we evaluate): an input projection, $N$ conditioned residual blocks each terminating in an SE channel-recalibration unit~\citep{hu2018se}, and an output projection, with the conditioning $c$ (defined in §\ref{sec:method:prelim}) injected via FiLM~\citep{perez2018film}. SE recalibrates channels using pooled spatial context, while FiLM injects prompt-conditioned feature modulation; these are practical design choices, not theoretically necessary components. The output projection, the FiLM heads, and a learnable null-prompt parameter are zero-initialized, so $f_\phi$ launches as the safe constant-zero predictor and the inference-time gradient is well-defined throughout early training.

$f_\phi$ is trained on clean ground truth latent pairs from the same dataset the backbone is trained on with an MSE-plus-cosine loss:
\begin{equation}
  \mathcal{L}_\phi \;=\; \mathrm{MSE}\big(f_\phi(z, c),\; \Delta_\tau z\big) \;+\; \alpha\,\Big(1 - \cos\angle\big(f_\phi(z, c),\; \Delta_\tau z\big)\Big),
  \qquad \alpha = 0.5,
  \label{eq:phi}
\end{equation}
that disentangles direction from magnitude at the spatial-field granularity. Two practicalities matter. \emph{(i) Diffusion-aligned input augmentation.} At inference $f_\phi$ is fed $\hat x_0$, which carries timestep-dependent residual noise rather than a clean latent; with probability $p_\mathrm{aug} = 0.5$ we therefore replace the input by $z + \sigma_{t'}\epsilon$ for $t'$ sampled uniformly from the scheduler timestep grid, matching the sampling-time input distribution without modifying the diffusion forward process. \emph{(ii) Classifier-free training.} With probability $p_\mathrm{drop} = 0.2$ the prompt is replaced by the learnable null embedding, so $f_\phi$ jointly learns conditional and unconditional behaviour and remains well-defined when no prompt is supplied at sampling time.

\boldparagraph{Why guidance on $\hat\epsilon$ rather than on $z_t$.}
We apply motion guidance to the noise prediction $\hat\epsilon$ rather than directly editing the latent $z_t$: editing $z_t$ desynchronizes the sampler trajectory from the noise estimate used by the next update, whereas editing $\hat\epsilon$ keeps the prior aligned with the diffusion update law. Appendix~\ref{app:guide_space} provides the full rationale and ablation evidence.

\boldparagraph{Guidance loss and inference rule.}
At sampling step $s$, after the standard CFG mix has produced $\hat\epsilon_\mathrm{CFG}$, we form the per-pair motion-consistency loss:
\begin{equation}
  \mathcal{L}_\mathrm{guide} \;=\; \frac{1}{T-\tau}\sum_{i=1}^{T-\tau}\big\|\hat x_0^{(i)} + f_\phi\big(\hat x_0^{(i)}, c\big) \;-\; \hat x_0^{(i+\tau)}\big\|_2^2,
  \label{eq:guide_loss}
\end{equation}
which asks $\hat x_0^{(i+\tau)}$ to be exactly one $\tau$-step motion increment away from $\hat x_0^{(i)}$ as predicted by $f_\phi$. We then apply guidance only after the first ${\lfloor(1 - \rho)\,S\rfloor}$ sampling steps have elapsed (default $\rho = {0.85}$, indexing $s$ from $0$ at the high-noise end): early steps are dominated by appearance assembly where $\hat x_0$ is too noisy for $f_\phi$ to be reliable, and gating guidance to the lower-noise window lets motion supervision act precisely when it can. The full inference rule is:
\begin{equation}
  \hat\epsilon_\mathrm{guided} \;=\; \hat\epsilon_\mathrm{CFG} \;-\; \lambda_\mathrm{guide}\cdot\mathbf{1}\big\{s \ge {\lfloor(1 - \rho)\,S\rfloor}\big\}\cdot\nabla_{\hat\epsilon}\mathcal{L}_\mathrm{guide}\big(\hat x_0(\hat\epsilon, z_t, t),\, c\big),
  \label{eq:guide}
\end{equation}
applied after the CFG mix so that the gradient is taken with respect to the prompt-consistent noise estimate. Eq.~\ref{eq:guide} adds a single forward-and-backward through the small CNN per sampling step; $f_\phi$ is frozen and no architectural change to the scheduler or backbone is required. We refer to this update as \textbf{Motion Prior Guidance}, denoted $\mathcal{G}_\mathrm{motion}$.

\subsection{Full Pipeline and Cost}
\label{sec:method:pipeline}

We instantiate \thename{} on top of the open-source CogVideoX~\citep{yang2024cogvideox} family. At training time, the diffusion backbone is fine-tuned with $\mathcal{L}_\mathrm{train}$ (Eq.~\ref{eq:train}) against a macro drift target read directly from each minibatch, while the micro motion-field predictor $f_\phi$ is trained on clean latent pairs from the same video data via Eq.~\ref{eq:phi}. Since $f_\phi$ depends only on clean latents and prompt conditioning, not on the current backbone state, the two objectives can be optimized separately or in parallel.

At inference time, the backbone architecture, input/output interface, and backbone parameter count are unchanged. The only added component is the frozen $\sim$10M-parameter predictor $f_\phi$, which contributes one lightweight forward-and-backward pass during the active guidance window of Eq.~\ref{eq:guide}. Thus \thename{} surfaces motion supervision from the same raw videos used for diffusion fine-tuning, without simulators, physical annotations, or foundation-model teachers, while keeping the pretrained video diffusion backbone itself intact.

\section{Experiments}
\label{sec:experiments}

\subsection{Experimental Setup}
\label{sec:exp:setup}

We evaluate on VideoPhy~\citep{bansal2024videophy}, VideoPhy2~\citep{bansal2025videophy2}, and VBench~\citep{huang2024vbench}. VideoPhy/VideoPhy2 report Semantic Adherence (SA) and Physical Commonsense (PC), while VBench measures general video quality and motion-related dimensions. PisaBench~\citep{li2025pisabench} provides reference-based object-motion metrics. We separately report an image-conditioned Physics-IQ evaluation, using its own matched control and protocol. We fine-tune CogVideoX~\citep{yang2024cogvideox} backbones on OpenVid~\citep{nan2024openvid} following VideoREPA's protocol~\citep{zhang2025videorepa}, without physics annotations, simulator rollouts, or external teacher signals. Full training and inference details are provided in Appendix~\ref{app:setup}.

\begin{table}[ht]
\centering
\caption{VideoPhy results. SA measures semantic adherence and PC measures physical commonsense. Bold and underline denote the best and second-best values within each comparison block.}
\resizebox{\linewidth}{!}
{
\begin{tabular}{lccccccccc}
\toprule
\multirow{2}{*}{Methods} & \multirow{2}{*}{Extra Supervision} & \multicolumn{2}{c}{Solid-Solid} & \multicolumn{2}{c}{Solid-Fluid} & \multicolumn{2}{c}{Fluid-Fluid} & \multicolumn{2}{c}{Overall} \\
\cmidrule(lr){3-4} \cmidrule(lr){5-6} \cmidrule(lr){7-8} \cmidrule(lr){9-10}
& & SA & PC & SA & PC & SA & PC & SA & PC\\
\midrule
CogVideoX-2B & - & 49.6 & 13.3 & 71.2 & 28.1 & 60.0 & 50.9 & 60.5 & 25.6 \\
DreamWorld-1.3B & VGGT, DINOv2 & \underline{54.5} & \textbf{24.5} & 48.6 & 25.4 & 60.1 & 32.7 & 52.9 & 26.2 \\
\textcolor{gray}{MoAlign-2B (paper)} & \textcolor{gray}{VideoMAE} & \textcolor{gray}{24.7} & \textcolor{gray}{31.7} & \textcolor{gray}{66.9} & \textcolor{gray}{40.7} & \textcolor{gray}{67.3} & \textcolor{gray}{56.4} & \textcolor{gray}{49.3} & \textcolor{gray}{39.4} \\
MoAlign-2B (reimpl.) & VideoMAE & 54.6 & \underline{18.3} & 73.5 & \underline{31.9} & \underline{66.2} & \underline{55.7} & \underline{64.5} & \underline{30.1} \\
VideoREPA-2B & VideoMAEv2 & 52.4 & 18.2 & \textbf{77.4} & \textbf{32.2} & 60.0 & 52.7 & 64.2 & 29.7 \\
LaMo-2B (Ours) & Self-supervised & \textbf{58.7} & 16.8 & \underline{74.7} & \textbf{32.2} & \textbf{69.1} & \textbf{67.3} & \textbf{67.2} & \textbf{31.4} \\
\midrule
CogVideoX-5B & - & \textbf{62.9} & 19.6 & 76.0 & 33.6 & 72.7 & 61.8 & 70.0 & 32.3 \\
PhyT2V-5B & o1-preview  & - & - & - & - & - & - & 61 & 37 \\
WISA-5B & Qwen2VL & - & - & - & - & - & - & 67 & 38 \\
PHANTOM-5B & V-JEPA2 & - & - & - & - & - & - & 47.5 & 37.9 \\
MoAlign-5B (reimpl.) & VideoMAE & \underline{62.5} & 26.3 & 79.6 & 38.4 & 78.0 & \underline{76.2} & \underline{72.2} & 39.4 \\
VideoREPA-5B & VideoMAEv2 & 58.0 & \textbf{28.0} & \textbf{82.9} & \underline{39.0} & \textbf{80.0} & 74.5 & 72.1 & \underline{40.1} \\
LaMo-5B (Ours) & Self-supervised & \textbf{62.9} & \underline{26.6} & \underline{80.8} & \textbf{41.1} & \underline{78.2} & \textbf{78.2} & \textbf{73.0} & \textbf{41.0} \\
\bottomrule
\end{tabular}
}
\label{tab:videophy}
\end{table}

\begin{table}[h]
\centering
\begin{minipage}[t]{0.36\linewidth}
  \centering
  \caption{Quantitative results on the VideoPhy2 benchmark.}
  \small
  \setlength{\tabcolsep}{5pt}
  \begin{tabular}{lcc}
    \toprule
    Method & SA & PC \\
    \midrule
    CogVideoX-2B & 21.0 & 68.0 \\
    PHANTOM-5B & 27.8 & 71.7 \\
    \textcolor{gray}{MoAlign-2B (paper)} & \textcolor{gray}{28.8} & \textcolor{gray}{75.0} \\
    MoAlign-2B (reimpl.) & \underline{24.6} & \underline{73.1} \\
    VideoREPA-2B & 21.0 & 72.5 \\
    \thename{}-2B (Ours) & \textbf{25.4} & \textbf{75.4} \\
    \bottomrule
  \end{tabular}
  \label{tab:videophy2}
\end{minipage}\hfill
\begin{minipage}[t]{0.55\linewidth}
  \centering
  \caption{Quantitative results on the PisaBench benchmark, aggregated over 421 videos.}
  \small
  \setlength{\tabcolsep}{8pt}
  \begin{tabular}{lccc}
    \toprule
    Method & L2 $\downarrow$ & CD $\downarrow$ & IoU $\uparrow$ \\
    \midrule
    CogVideoX-5B-I2V & \underline{0.1180} & \underline{0.3017} & \underline{0.1481} \\
    Cosmos-Predict2.5 & 0.1373 & 0.3871 & 0.1467 \\
    Cosmos3-Nano & 0.1250 & 0.3437 & 0.1361 \\
    LTX-Video & 0.1181 & 0.3071 & 0.1017 \\
    OpenSoraV2 & 0.1293 & 0.3310 & 0.0846 \\
    \thename{}-5B (Ours) & \textbf{0.1113} & \textbf{0.2888} & \textbf{0.1590} \\
    \bottomrule
  \end{tabular}
  \label{tab:pisabench}
\end{minipage}
\end{table}

\subsection{Main Results}
\label{sec:exp:main}

\boldparagraph{VideoPhy.}
Table~\ref{tab:videophy} evaluates whether \thename{} improves physical fidelity without sacrificing semantic adherence. Relative to CogVideoX-2B, \thename{}-2B improves the overall score from $60.5$/$25.6$ to $67.2$/$31.4$ SA/PC, a gain of $+6.7$ SA and $+5.8$ PC. It also improves over the reproduced 2B physics-aware baselines on the overall SA/PC pair, while using only self-supervised video signals. At the larger scale, \thename{}-5B improves CogVideoX-5B from $70.0$/$32.3$ to $73.0$/$41.0$, and is slightly higher than VideoREPA-5B on both overall axes ($72.1$/$40.1$). The largest PC gains appear in the \emph{Fluid--Fluid} category, where \thename{} reaches $67.3$ PC at 2B and $78.2$ PC at 5B. This category involves less rigid, harder-to-script dynamics, so the improvement is consistent with our goal of extracting motion regularities from ordinary video rather than relying on hand-specified physics supervision.

\boldparagraph{VideoPhy2.}
On the more challenging VideoPhy2 benchmark (Table~\ref{tab:videophy2}), which targets a broader and harder distribution of physics-laden prompts, \thename{}-2B reaches $25.4$ SA / $75.4$ PC, improving over the CogVideoX-2B baseline by $+4.4$ SA / $+7.4$ PC and over the strongest prior 2B competitor, VideoREPA-2B ($21.0$/$72.5$), by $+4.4$ SA / $+2.9$ PC. Notably, \thename{}-2B also surpasses PHANTOM-5B ($27.8$/$71.7$) on PC despite being less than half its size, while approaching its SA score. These results indicate that the gains observed on VideoPhy generalize to a markedly more demanding and diverse distribution of physical phenomena.

\boldparagraph{PisaBench.}
On the 421-video evaluation in Table~\ref{tab:pisabench}, \thename{}-5B achieves the lowest trajectory L2 and CD and the highest IoU among the evaluated methods. These results show its strong performance on PisaBench relative to existing video generation methods.

\begin{table*}[ht]
\centering
\caption{Results on VBench. \thename{} improves all reported dimensions over CogVideoX, indicating that the motion prior preserves video quality while strengthening motion and spatial consistency.}
\resizebox{\linewidth}{!}
{
\begin{tabular}{l|ccccccc|ccc}
\toprule
Method & \makecell{Motion\\Smooth.} & \makecell{Multi.\\Obj.} & \makecell{Obj.\\Class} & \makecell{Overall\\Consist.} & Scene & \makecell{Spatial\\Relation.} & \makecell{Temp.\\Flicker.} & \makecell{Quality\\Score} & \makecell{Semantic\\Score} & \makecell{Total\\Score} \\
\midrule
CogVideoX & 97.6 & 50.4 & 78.7 & 25.0 & 40.3 & 52.3 & 97.3 & 80.5 & 68.7 & 78.2 \\
LaMo (Ours) & \textbf{98.2} & \textbf{51.6} & \textbf{82.0} & \textbf{25.7} & \textbf{42.6} & \textbf{62.2} & \textbf{98.4} & \textbf{81.9} & \textbf{70.7} & \textbf{79.6} \\
\bottomrule
\end{tabular}
}
\label{tab:vbench}
\end{table*}

\begin{table}[t]
\centering
\begin{minipage}[t]{0.47\linewidth}
  \centering
  \caption{Key component ablation on VideoPhy.}
  \small
  \renewcommand{\arraystretch}{1.35}
  \setlength{\tabcolsep}{5pt}
  \begin{tabular}{lcc}
    \toprule
    Component & SA & PC \\
    \midrule
    Baseline & 64.8 & 35.2 \\
    + $\mathcal{L}_\mathrm{drift}$ only & \underline{71.8} & \underline{39.0} \\
    + $\mathcal{G}_\mathrm{motion}$ only & 68.9 & 38.4 \\
    \thename{} (full) & \textbf{73.0} & \textbf{41.0} \\
    \bottomrule
  \end{tabular}
  \renewcommand{\arraystretch}{1}
  \label{tab:ablation}
\end{minipage}\hfill
\begin{minipage}[t]{0.47\linewidth}
  \centering
  \caption{Design-choice ablation on VideoPhy.}
  \small
  \renewcommand{\arraystretch}{0.7}
  \setlength{\tabcolsep}{6pt}
  \begin{tabular}{lcc}
    \toprule
    Variant & SA & PC \\
    \midrule
    Dense motion loss & 64.2 & 34.6 \\
    Raw L2 motion loss & 67.3 & 36.1 \\
    Adj-frame lag ($\tau{=}1$) & \underline{68.0} & 37.5 \\
    \midrule
    Direct latent edit ($z_t$) & 62.8 & 31.6 \\
    No predictor aug. & 66.4 & \underline{37.6} \\
    \midrule
    \thename{} (Ours) & \textbf{73.0} & \textbf{41.0} \\
    \bottomrule
  \end{tabular}
  \label{tab:design_ablation}
\end{minipage}
\end{table}

\boldparagraph{VBench.}
Table~\ref{tab:vbench} checks whether the physics-oriented objective harms general generation quality. \thename{} improves the CogVideoX Total Score from $78.2$ to $79.6$, with gains in both Quality ($80.5\rightarrow81.9$) and Semantic ($68.7\rightarrow70.7$). The largest motion-related change is Spatial Relationship ($52.3\rightarrow62.2$), while Motion Smoothness and Temporal Flickering also increase slightly. Since all reported VBench dimensions move upward in this comparison, the results suggest that the latent motion prior improves physical and spatial consistency without an observed trade-off in the measured general-quality metrics.

\subsection{Ablation Study}
\label{sec:exp:ablation}

\boldparagraph{Effect of each component.}
Table~\ref{tab:ablation} ablates the two core components of \thename{} on VideoPhy. Starting from the baseline at $64.8$ SA / $35.2$ PC, the training-time Motion Drift Loss alone improves to $71.8$/$39.0$ ($+7.0$/$+3.8$), showing that a macro latent-change target already improves physical fidelity without modifying sampling. The inference-time Motion Prior Guidance alone reaches $68.9$/$38.4$ ($+4.1$/$+3.2$), indicating that the frozen motion-field predictor also contributes useful spatial motion structure. Combining both readouts gives the best result, $73.0$/$41.0$, higher than either single component on both SA and PC; this supports our stage-matching argument that macro drift anchors the global latent-change rate during training, while micro guidance allocates that change spatially during sampling.

\boldparagraph{Validating individual design choices.}
Table~\ref{tab:design_ablation} validates the main design choices by replacing each with the corresponding alternative. On the drift side, dense per-pixel supervision ($64.2$/$34.6$), raw L2 ($67.3$/$36.1$), and adjacent-frame lag $\tau{=}1$ ($68.0$/$37.5$) all trail the full method ($73.0$/$41.0$), supporting the channel-mean target, scale-normalized loss, and $\tau{=}2$ lag. On the guidance side, direct latent editing of $z_t$ drops to $62.8$/$31.6$, below the baseline, while removing predictor augmentation gives $66.4$/$37.6$; these results support noise-space guidance and diffusion-aligned predictor training rather than post-hoc alternatives.

\subsection{Qualitative Analysis}
\label{sec:exp:qualitative}
\begin{figure}[t]
  \centering
  \includegraphics[width=0.99\linewidth]{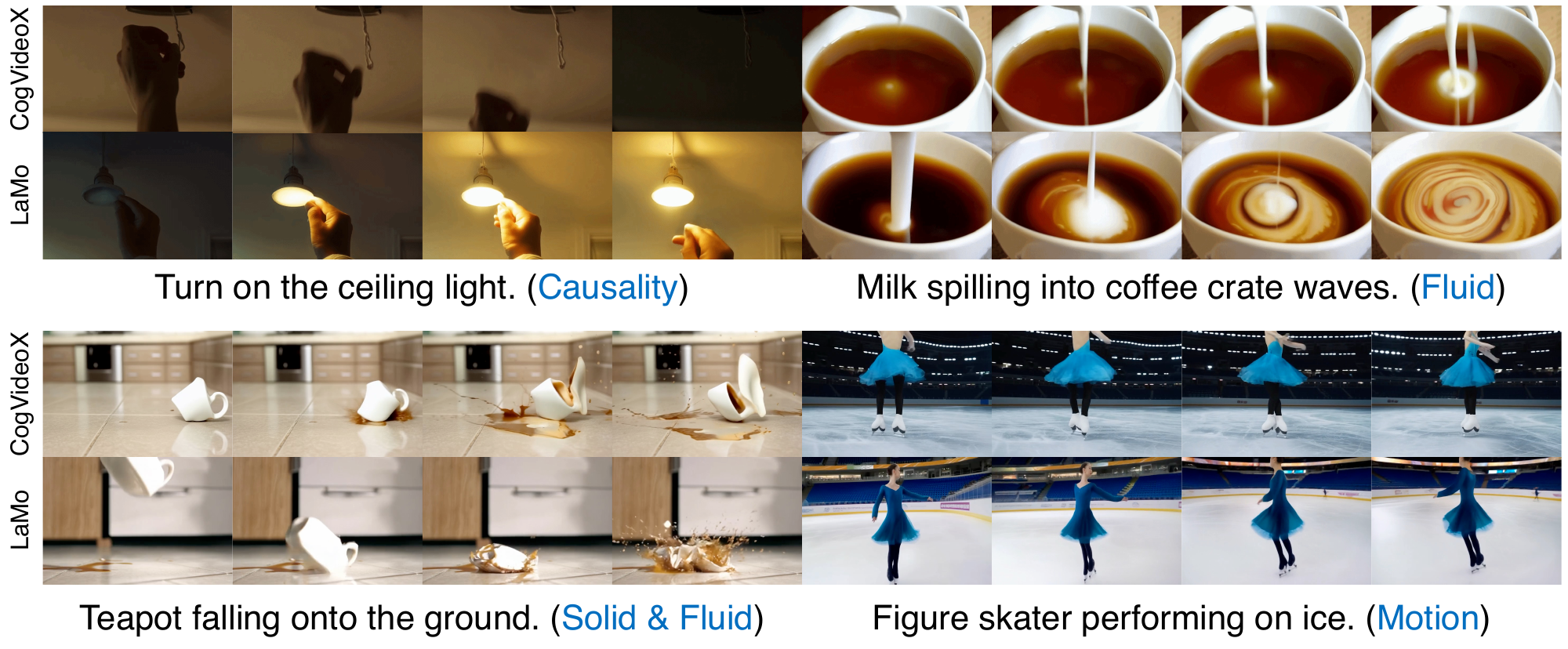}
\caption{\textbf{Qualitative comparisons on physics-heavy prompts.} \thename{} produces more coherent motion and more consistent relationships between actions and their effects.}
  \label{fig:qualitative}
\end{figure}

\boldparagraph{Qualitative comparisons.}
Figure~\ref{fig:qualitative} compares generated sequences across causal state changes, fluid dynamics, object-fluid impacts, and articulated human motion. CogVideoX produces plausible individual frames, but its temporal dynamics often fail to reflect the prompted events: switching on the light does not consistently change illumination, milk entering coffee shows limited waves or mixing, teapot contact and spilling are unstable, and the skater's legs become tangled across frames. In contrast, \thename{} better links actions to their consequences, producing consistent illumination, coherent waves and mixing, a localized splash and spill, and continuous skating motion with stable body structure.

\begin{figure}[t]
  \centering
  \includegraphics[width=0.99\linewidth]{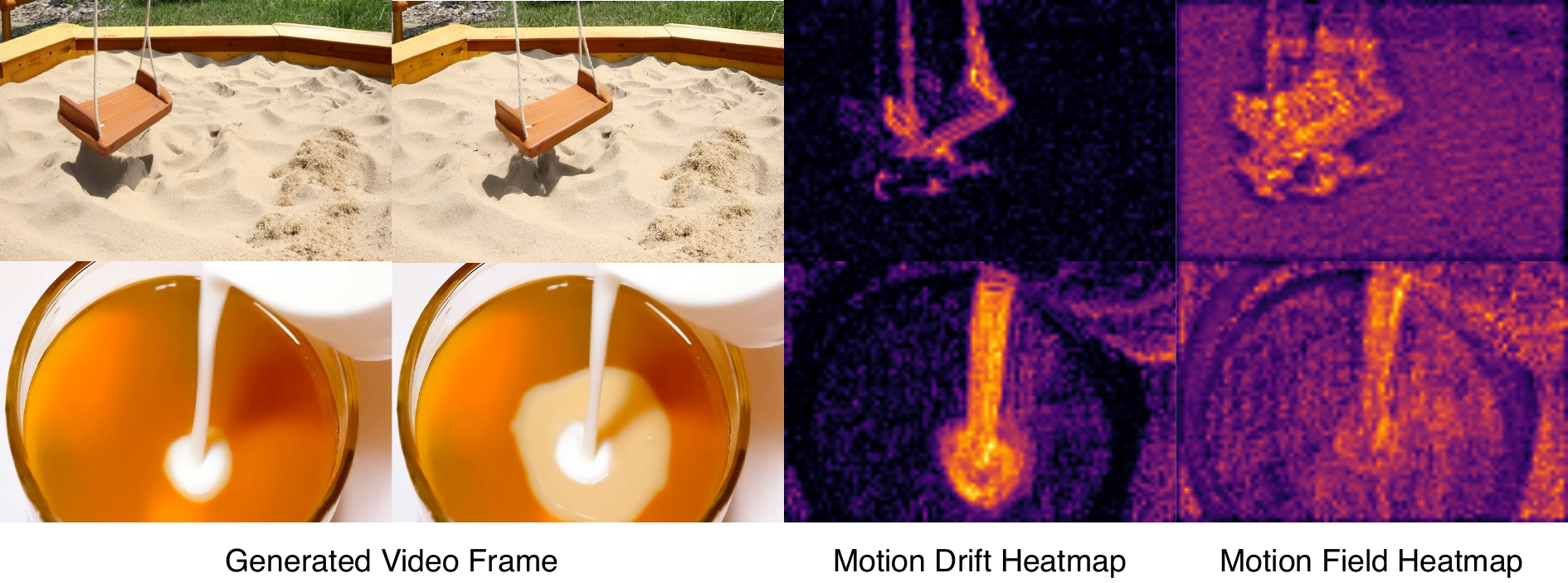}
\caption{\textbf{Interpretability of \thename{}'s latent motion readouts.} For each clip, we show sampled frames, a Motion Drift Heatmap, and a Motion Field Heatmap. The drift heatmap visualizes the macro latent-change direction supervised by $\mathcal{L}_\mathrm{drift}$; the field heatmap shows the response of $f_\phi$ used in $\mathcal{G}_\mathrm{motion}$. Both focus on physically active regions. Appendix~\ref{app:heatmap} describes the visualization.}
  \label{fig:interpretability}
\end{figure}

\boldparagraph{Heatmap interpretability.}
Figure~\ref{fig:interpretability} shows that the macro motion drift and micro motion field heatmaps concentrate on active regions, such as the moving swing and milk stream, rather than spreading uniformly over the scene. This localization supports the qualitative observation that \thename{} improves event-level dynamics rather than merely changing frame appearance. Appendix~\ref{app:heatmap} details the heatmap computation.

\section{Conclusions and Limitations}
\label{sec:conclusion}
\boldparagraph{Conclusions.}
We presented \thename{}, a self-supervised framework that extracts a latent motion prior from ordinary video and exposes it as a training-time Motion Drift Loss and sampling-time Motion Prior Guidance. Without simulators, teacher models, or physics annotations, \thename{} improves physical fidelity on VideoPhy, VideoPhy2, and PisaBench while preserving general generation quality on VBench.

\boldparagraph{Limitations.}
\thename{} encourages plausible latent motion but is not a constraint-satisfying simulator. Its gains are bounded by the coverage of the video data and by the granularity of VideoPhy-style model-judged metrics, which do not isolate contact, deformation, fluids, conservation, or long-horizon stability. Future work should pair latent motion priors with targeted physical measurements and integrate them more directly into large-scale pretraining.

Latent differences are compressed feature changes, not explicit displacement or trajectory fields. They can mix object motion with camera motion, illumination, or occlusion; VAE compression may hide fine-scale contact, and spatial averaging can attenuate localized or opposing motion. The spatially resolved micro readout mitigates the last issue but does not remove these limitations. LaMo is complementary to explicit correspondence cues such as optical flow. The reference-based evaluations cover specific motion settings and do not replace a controlled human study or establish long-horizon physical correctness.

We use latent differences to supervise the generator in its native state space, but have not established that they are cleaner than RGB differences through a controlled pixel-target comparison. VAE compression may suppress some variation while retaining appearance and camera changes.

\boldparagraph{Broader impacts.}
More physically realistic generated videos could support simulation and synthetic-data creation. However, generated videos may still contain physically implausible motion or interactions, limiting their reliability in these applications. Evaluation and release should account for misleading uses of generated content as well as failures in physical fidelity.

\bibliographystyle{ACM-Reference-Format}
\bibliography{main}

\clearpage
\appendix
\setlength{\intextsep}{6pt plus 2pt minus 2pt}
\section{Heatmap Computation for Figure~\ref{fig:interpretability}}
\label{app:heatmap}

We compute the two heatmaps in Figure~\ref{fig:interpretability} on the same generated latent trajectory so that the macro and micro readouts are visualized at the same physical event. Let $z \in \mathbb{R}^{1 \times T \times C \times H \times W}$ denote the clean video latent produced for a prompt $c$, and let $\tau=2$ be the same temporal lag used in training. For each valid latent frame index $t$, we compute
\begin{equation}
  \Delta_t = z^{(t+\tau)} - z^{(t)}, \qquad
  b_t = \frac{1}{HW}\sum_{h,w}\Delta_t(:,h,w) \in \mathbb{R}^{C},
\end{equation}
where $b_t$ is the empirical macro drift vector for that frame pair. We choose the visualization index
\begin{equation}
  t^\star = \arg\max_t \|b_t\|_2,
\end{equation}
which selects the pair with the largest spatially averaged latent change. The RGB frame shown in the first column is the decoded video frame corresponding to $t^\star$ under the VAE temporal compression.

\boldparagraph{Motion Drift Heatmap.}
The Motion Drift Heatmap visualizes how strongly each spatial cell's latent change aligns with the macro drift direction. For the selected frame pair, we compute
\begin{equation}
  R_\mathrm{drift}(h,w)
  =
  \frac{\left|\left\langle \Delta_{t^\star}(:,h,w),\, b_{t^\star}\right\rangle\right|}
       {\|b_{t^\star}\|_2 + \varepsilon}.
\end{equation}
This quantity is parameter-free: it is the per-spatial-cell projection of the observed latent difference onto the channel-mean drift direction supervised by $\mathcal{L}_\mathrm{drift}$. Taking the absolute value visualizes the magnitude of aligned latent change regardless of the sign of the projection.

\boldparagraph{Motion Field Heatmap.}
The Motion Field Heatmap visualizes the spatial response of the learned predictor $f_\phi$. A direct norm $\|f_\phi(z^{(t^\star)}, c)(:,h,w)\|_2$ is often dense because a CNN predictor can produce a non-zero default response even in static background regions. We therefore subtract a no-motion baseline. Specifically, we form a temporally averaged static latent
\begin{equation}
  z_\mathrm{static} = \frac{1}{T}\sum_{t=1}^{T} z^{(t)},
\end{equation}
and compute
\begin{equation}
  R_\mathrm{field}(h,w)
  =
  \left\|
  f_\phi(z^{(t^\star)}, c)(:,h,w)
  -
  f_\phi(z_\mathrm{static}, c)(:,h,w)
  \right\|_2.
\end{equation}
This time-averaged baseline retains aggregate scene context and reduces the predictor's dense background response. The subtraction is used only for visualization, not for training or sampling.

\section{Experimental Setup Details}
\label{app:setup}

\boldparagraph{Benchmarks.}
VideoPhy~\citep{bansal2024videophy} contains $344$ prompts spanning three interaction categories: \emph{Solid--Solid}, \emph{Solid--Fluid}, and \emph{Fluid--Fluid}. VideoPhy and VideoPhy2~\citep{bansal2025videophy2} report Semantic Adherence (SA), measuring whether the generated content matches the prompt, and Physical Commonsense (PC), measuring whether the generated dynamics obey elementary physical regularities. PisaBench~\citep{li2025pisabench} measures image-conditioned object motion using trajectory L2, Chamfer Distance, and IoU. We also evaluate image-conditioned continuation on Physics-IQ~\citep{motamed2026physics}; these protocols differ from the text-to-video VideoPhy tests. VBench~\citep{huang2024vbench} evaluates general video generation quality along sixteen dimensions; in the main paper we report Quality, Semantic, and Total scores together with seven motion- and consistency-related sub-dimensions: Motion Smoothness, Temporal Flickering, Multiple Objects, Object Class, Overall Consistency, Scene, and Spatial Relationship.

\boldparagraph{Training data.}
Following VideoREPA~\citep{zhang2025videorepa}, we fine-tune all backbones on a clean subset of OpenVid~\citep{nan2024openvid}. \thename{}-2B is trained on $32$k clips and \thename{}-5B on $64$k clips, both filtered to $480$p and $49$ frames per clip. No physics annotations, simulator rollouts, or external teacher signals are used.

\begin{figure}[ht]
  \centering
  \includegraphics[width=0.99\linewidth]{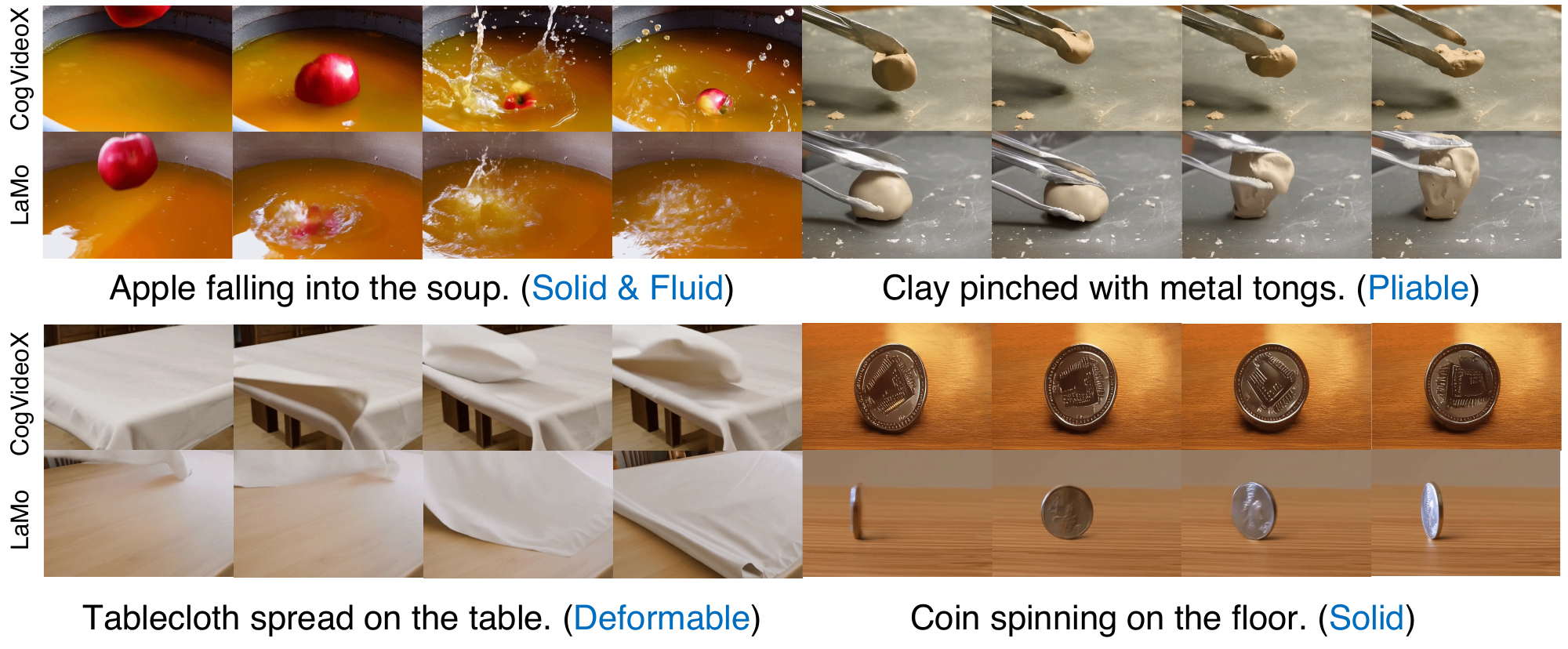}
  \caption{\textbf{Additional qualitative results on physics-heavy prompts.} The sampled frames compare liquid impact, cloth spreading, contact deformation, and rotational continuity.}
  \label{fig:vis_supp}
\end{figure}

\boldparagraph{Implementation details.}
We instantiate \thename{} on CogVideoX-2B and CogVideoX-5B~\citep{yang2024cogvideox}. The 2B model is fully fine-tuned, while the 5B model uses LoRA~\citep{hu2022lora} with rank $128$ and alpha $64$. All experiments use $16$ A100 GPUs, global batch size $32$, and $2$k optimizer steps. We set the latent-change lag to $\tau{=}2$ and the Motion Drift Loss weight to $\lambda_\mathrm{drift}{=}0.4$. The motion-field predictor $f_\phi$ has $\sim$10M parameters and is trained with diffusion-aligned input augmentation probability $p_\mathrm{aug}{=}0.5$ and classifier-free dropout probability $p_\mathrm{drop}{=}0.2$. At inference, Motion Prior Guidance uses $\lambda_\mathrm{guide}{=}25.0$ and active-window ratio $\rho{=}0.85$.

\section{Additional Qualitative Comparisons}
\label{app:vis}

To further substantiate the qualitative observations in the main paper, Figure~\ref{fig:vis_supp} shows four additional prompts that stress different physical regimes: solid--fluid interaction, deformable cloth, pliable material, and rigid-body angular motion. For each prompt, we generate videos with the CogVideoX baseline and with \thename{} under the same prompt, seed, and sampler configuration, and visualize representative frames sampled at matched temporal positions. Across these examples, \thename{} shows more coherent event progression in the sampled frames.

\boldparagraph{Apple falling into soup (solid--fluid).}
The prompt describes an apple dropping into a bowl of soup. Both sampled sequences show an impact, while \thename{} depicts more pronounced local liquid displacement and a deeper entry in the shown frames. These images illustrate the impact sequence; they do not establish the apple's eventual buoyant state.

\boldparagraph{Tablecloth spread on the table (deformable).}
The prompt describes a tablecloth being spread onto a table. The baseline frames show the cloth mostly laid out from the outset, while \thename{} shows fabric above the table followed by spreading onto the surface, better matching the event order in the prompt.

\boldparagraph{Clay pinched with metal tongs (pliable).}
The prompt depicts tongs pinching a clay block. In the sampled frames, the baseline shows limited local deformation; \thename{} shows a clearer indentation near contact and a change in shape. This supports a qualitative comparison of contact deformation, but does not establish the material's plastic response.

\boldparagraph{Coin spinning on the floor (solid).}
The prompt depicts a coin spinning on a flat surface. Across the sampled frames, the baseline changes the coin's orientation irregularly, while \thename{} shows a more continuous progression of orientation and tilt. These frames support a qualitative comparison of rotational continuity, not a measured claim about angular velocity or friction.

\begin{figure}[ht]
  \centering
  \includegraphics[width=0.99\linewidth]{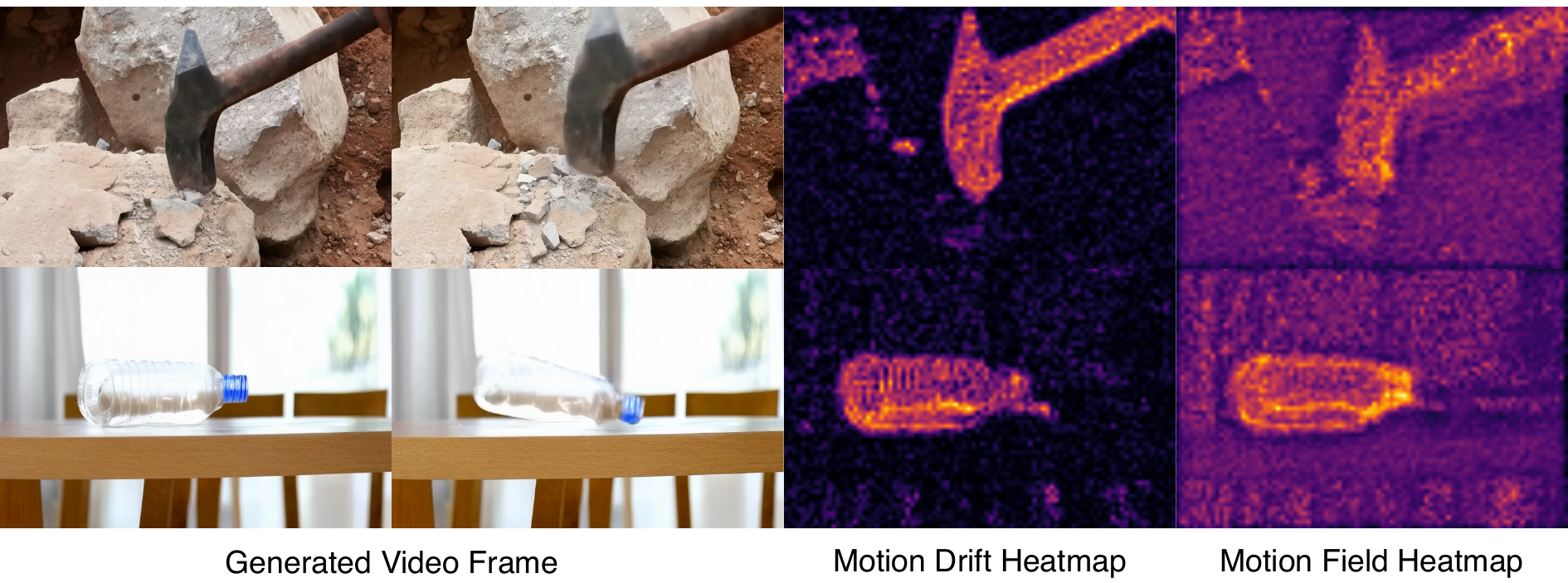}
  \caption{\textbf{Additional interpretability visualizations of \thename{}'s latent motion readouts.} Each row shows a sampled RGB frame, Motion Drift Heatmap, and Motion Field Heatmap (Appendix~\ref{app:heatmap}). Top: \emph{hammer striking a stone}. Bottom: \emph{water bottle rolling on a table}. Both heatmaps emphasize regions associated with the depicted motion.}
  \label{fig:interp_supp}
\end{figure}

\boldparagraph{Summary.}
Across these four prompts, the sampled LaMo frames show more coherent impact, cloth spreading, contact deformation, and rotation. These visual examples complement the quantitative results; sampled frames alone cannot establish complete physical trajectories.

\section{Additional Interpretability Visualizations}
\label{app:interp}

We further visualize \thename{}'s two latent motion readouts on a hammer-strike clip and a rolling-bottle clip in Figure~\ref{fig:interp_supp}. Following Appendix~\ref{app:heatmap}, we select the latent frame pair with the strongest macro drift and show the corresponding decoded frame, the projected latent-change heatmap, and the predictor-response heatmap. The time-averaged baseline reduces dense background response only in this visualization.

\boldparagraph{Hammer striking a stone.}
The sampled frame depicts a hammer striking a stone. Both heatmaps respond near the hammer head and contact region, with a more localized field response. This links the readouts to the visible action without measuring impact forces.

\boldparagraph{Water bottle rolling on a table.}
In the rolling-bottle clip, both heatmaps respond on the bottle rather than the background. The field heatmap resolves more of its outline, consistent with a spatially resolved readout; the images do not quantify rolling speed.

\boldparagraph{Discussion.}
In both examples, the visualized responses are strongest near the moving objects and contact regions rather than the background. These heatmaps help interpret the two readouts, but do not by themselves establish physical correctness.

\section{Additional Details on Guidance Space}
\label{app:guide_space}

A natural alternative to Eq.~\ref{eq:guide} is to apply the motion-consistency gradient directly to the latent $z_t$, in the spirit of latent-space score editing. We avoid this choice because the diffusion sampler defines its latent update through $\hat\epsilon$ and the noise schedule. An external edit on $z_t$ therefore acts at the wrong scale for the current step, desynchronizing the latent trajectory chosen by the scheduler from the noise estimate expected by the next step. This mismatch can compound across sampling and visibly degrade sample quality.

\begin{table}[ht]
\centering
\caption{Matched evaluation on Wan2.2-TI2V-5B.}
\label{tab:wan}
\small
\begin{tabular}{lccccc}
\toprule
\multirow{2}{*}{Method} & \multicolumn{2}{c}{VideoPhy} & \multicolumn{3}{c}{VideoPhy2} \\
\cmidrule(lr){2-3}\cmidrule(lr){4-6}
& SA $\uparrow$ & PC $\uparrow$ & SA $\uparrow$ & PC $\uparrow$ & Joint $\uparrow$ \\
\midrule
Wan2.2-5B baseline & 65.70 & \textbf{30.52} & 25.76 & 68.47 & 23.05 \\
Wan2.2-5B + \thename{} & \textbf{65.99} & 29.65 & \textbf{26.61} & \textbf{70.17} & \textbf{23.22} \\
\bottomrule
\end{tabular}
\end{table}

A correction specified on $\hat x_0$ can be mapped algebraically to the scheduler's native prediction; this affine conversion does not itself require an iterative inversion or imply accumulated inversion error. We apply the motion-consistency gradient directly to the native prediction so that the scheduler interface and timestep scaling are explicit. This follows the same principle as classifier-free guidance, which corrects the model prediction rather than editing the sampled trajectory. This keeps the motion prior aligned with the diffusion update law and gives a built-in time-dependent attenuation through $\hat x_0$: by the chain rule, the guidance prefactor scales as $\sigma_t/\sqrt{\bar\alpha_t}$ for $\epsilon$-prediction or $\sigma_t$ for $v$-prediction, so the correction naturally shrinks as $\bar\alpha_t \to 1$.

The ablation in Table~\ref{tab:design_ablation} supports this design choice. The direct-$z_t$ variant drops to $62.8$ SA / $31.6$ PC, below the unaugmented backbone, consistent with the risk of scheduler mismatch, although this ablation does not isolate that mechanism.

\section{Physics-IQ Evaluation}
\label{app:reference_eval}

Physics-IQ requires continuation from a fixed initial frame. We therefore trained matched image-to-video CogVideoX-5B control and LaMo adaptations on the same OpenVid data, with identical adapter initialization, optimizer, and $8$-GPU, $2{,}000$-step budget. Only the LaMo arm adds the motion-field predictor and motion objectives. This shorter adaptation budget differs from the main CogVideoX-5B recipe.

We evaluate all 38 scenarios of the Original Physics-IQ Solid Mechanics category from three fixed camera views (114 conditions). The matched LaMo adaptation scores 38.56 versus 36.31 for its CogVideoX-5B control, a 2.25-point improvement on this category.

\section{Wan2.2 Flow-Matching Transfer}
\label{app:wan}

To probe transfer beyond CogVideoX, we ported LaMo to Wan2.2-TI2V-5B, which has a different 48-channel causal spatiotemporal VAE and a flow-matching sampler. We re-encoded OpenVid with Wan's VAE, recomputed latent-motion targets, trained a 48-channel motion-field predictor, and fine-tuned matched baseline and LaMo models using the same data, optimizer, seed, $2{,}000$-step budget, and LoRA rank/alpha of $128/64$. The target is VAE-specific; plug-and-play refers to retaining the target backbone architecture and I/O, not zero-shot transfer of a previously trained predictor.

For the linear-flow convention in this implementation, $x_t=(1-\sigma_t)x_0+\sigma_t\epsilon$, $u=\epsilon-x_0$, and the model-side clean-latent estimate is $\hat x_0=x_t-\sigma_t\hat u$. The drift loss remains on $\hat x_0$; at sampling, guidance acts on the native predicted field $\hat u$ through $\nabla_{\hat u}\mathcal{L}_\mathrm{guide}(\hat x_0,c)$. Since $\partial\hat x_0/\partial\hat u=-\sigma_t I$, the corresponding correction to $\hat x_0$ is proportional to $-\sigma_t^2\nabla_{\hat x_0}\mathcal{L}_\mathrm{guide}$ and attenuates as $\sigma_t\to0$. This affine conversion introduces no iterative inversion.

Table~\ref{tab:wan} shows small improvements on VideoPhy SA and all three VideoPhy2 scores, alongside a decline in VideoPhy PC. It provides initial evidence of transfer to another VAE and denoising parameterization, rather than a universal improvement guarantee.

Wan2.2 still uses a causal spatiotemporal VAE, so this experiment does not establish transfer to a framewise 2D VAE. Because $\tau$ is measured in latent frames, a different VAE temporal stride changes the physical duration of the target difference and may require retuning $\tau$.

\section{Inference Cost and Sensitivity}
\label{app:cost_sensitivity}

\boldparagraph{Inference cost.}
We measured end-to-end generation on one A100-80GB with bfloat16, $720\times480$ output, 49 frames, and 50 denoising steps, excluding model loading. At strength 25, guidance runs for the final 43 steps. Full LaMo adds 4.656\% runtime and 0.248\% peak memory relative to the public CogVideoX-2B run. On the matched predictor path, turning on guidance adds 1.311\% runtime and no additional backbone forward pass. The 43-step window uses the default $\rho=0.85$. The adapter-only and guidance-off rows use different inference call paths, so their times are not a component timing comparison.

\begin{table}[t]
\centering
\caption{Inference metrics on one A100-80GB. Mean $\pm$ SD over eight matched trials.}
\label{tab:runtime}
\small
\begin{tabular}{lcccc}
\toprule
\multirow{2}{*}{Condition} & E2E time & Peak mem. & Runtime & Memory \\
& (s/video) $\downarrow$ & (GB) $\downarrow$ & increase (\%) $\downarrow$ & increase (\%) $\downarrow$ \\
\midrule
Public CogVideoX-2B & $95.308 \pm 1.407$ & $8.929$ & $0$ & $0$ \\
\thename{} adapter only & $99.780 \pm 0.825$ & $8.929$ & $+4.708$ & $0$ \\
\thename{} predictor, guidance off & $98.443 \pm 0.917$ & $8.951$ & $+3.312$ & $+0.248$ \\
Full \thename{} & $99.725 \pm 0.914$ & $8.951$ & $+4.656$ & $+0.248$ \\
\bottomrule
\end{tabular}
\end{table}

\begin{table}[H]
\centering
\caption{Guidance strength and active-window sensitivity on VideoPhy2.}
\label{tab:guidance_sensitivity}
\small
\begin{tabular}{cccc}
\toprule
$\lambda_\mathrm{guide}$ & $\rho$ & SA $\uparrow$ & PC $\uparrow$ \\
\midrule
15 & 0.80 & 25.25 & 74.41 \\
20 & 0.80 & 24.75 & 74.92 \\
25 & 0.80 & 23.56 & 74.07 \\
25 & 0.85 & \textbf{25.42} & \textbf{75.42} \\
\bottomrule
\end{tabular}
\end{table}

\boldparagraph{Guidance sensitivity.}
Table~\ref{tab:guidance_sensitivity} reports the available VideoPhy2 sweep. Across the tested settings, SA ranges from 23.56 to 25.42 and PC from 74.07 to 75.42. The default $(\lambda_\mathrm{guide},\rho)=(25,0.85)$ has the highest scores among these tested settings; this limited sweep does not establish a global optimum.

\begin{table}[H]
\centering
\caption{Joint SE and FiLM ablation in the micro predictor on VideoPhy.}
\label{tab:se_film_ablation}
\small
\begin{tabular}{lcc}
\toprule
Micro predictor & SA $\uparrow$ & PC $\uparrow$ \\
\midrule
Vanilla CNN & 62.79 & 28.49 \\
CNN + SE + FiLM & \textbf{63.37} & \textbf{30.23} \\
\bottomrule
\end{tabular}
\end{table}

\boldparagraph{Micro-predictor architecture.}
The matched $8$-GPU, $2{,}000$-step VideoPhy ablation in Table~\ref{tab:se_film_ablation} compares a vanilla CNN with a predictor containing both SE and FiLM. The joint addition improves SA by 0.58 and PC by 1.74 points; this experiment does not isolate the modules individually.

\end{document}